%% file: main.tex
\documentclass{article}

\usepackage[utf8]{inputenc}
\usepackage[margin=1.5in]{geometry}

\usepackage{natbib}
\usepackage{graphicx}
\usepackage{tabularx}
\usepackage{placeins}
\usepackage[toc,page]{appendix}

\usepackage[utf8]{inputenc} % allow utf-8 input
\usepackage[T1]{fontenc}    % use 8-bit T1 fonts
\usepackage{hyperref}       % hyperlinks
\hypersetup{ %
    pdfborder=0 0 0,
    pdfpagemode=UseNone,
    colorlinks=true,
    linkcolor=blue,
    citecolor=blue,
    filecolor=blue,
    urlcolor=blue,
    pdfview=FitH}
    
\usepackage{url}            % simple URL typesetting
\usepackage{booktabs}       % professional-quality tables
\usepackage{amsfonts}       % blackboard math symbols
\usepackage{nicefrac}       % compact symbols for 1/2, etc.
\usepackage{microtype}      % microtypography
\usepackage{color}
\usepackage[dvipsnames]{xcolor}
\usepackage{amsmath}

\usepackage[font=small,labelfont=bf]{caption}
\usepackage[font=small,labelfont=bf]{subcaption}
\usepackage{mathtools}
\usepackage{verbatim}
\usepackage{algorithm}
\usepackage[noend]{algpseudocode}
\usepackage{multirow}
\usepackage{arydshln}
\usepackage{rotating}
\usepackage{array}
\usepackage{mfirstuc}
\usepackage{autobreak}

\input{macros.tex}
\DeclarePairedDelimiterX{\divx}[2]{\big(}{\big)}{%
  #1\;\delimsize\|\;#2%
}

\newcommand{\kldiv}{D_{KL}\divx}

\newcommand{\ourawesomemodel}{\emph{MONet}\xspace}
\newcommand{\oir}{\emph{Objects Room}\xspace}
\newcommand{\multidsprites}{\emph{Multi-dSprites}\xspace}
\newcommand{\clevr}{\emph{CLEVR}\xspace}
\newcommand{\allinone}{\emph{all-in-one}\xspace}
\newcommand{\elementmasks}{\emph{element masks}\xspace}
\newcommand{\wrongelementmasks}{\emph{wrong element masks}\xspace}

\title{MONet: Unsupervised Scene Decomposition and Representation}
\author{{
  Christopher P. Burgess, Loic Matthey, Nicholas Watters, }\\
  {Rishabh Kabra, Irina Higgins, Matt Botvinick, Alexander Lerchner} \vspace{5pt}
  \\
  DeepMind\\
  London, United Kingdom\vspace{5pt}
  \\
  \texttt{\{cpburgess, lmatthey, nwatters,} \\
  \texttt{rkabra, irinah, botvinick, lerchner\}@google.com}\vspace{-10pt}
}
\date{}

\begin{document}

\maketitle

\begin{abstract}
The ability to decompose scenes in terms of abstract building blocks is crucial for general intelligence. Where those basic building blocks share meaningful properties, interactions and other regularities across scenes, such decompositions can simplify reasoning and facilitate imagination of novel scenarios. In particular, representing perceptual observations in terms of entities should improve data efficiency and transfer performance on a wide range of tasks. Thus we need models capable of discovering useful decompositions of scenes by identifying units with such regularities and representing them in a common format. To address this problem, we have developed the Multi-Object Network (MONet). In this model, a VAE is trained end-to-end together with a recurrent attention network -- in a purely unsupervised manner -- to provide attention masks around, and reconstructions of, regions of images. We show that this model is capable of learning to decompose and represent challenging 3D scenes into semantically meaningful components, such as objects and background elements.
\end{abstract}

\input{introduction.tex}

\input{model.tex}

\input{results.tex}

\input{related_work.tex}

\input{conclusion.tex}

\subsubsection*{Acknowledgments}

We thank Ari Morcos, Daniel Zoran and Luis Piloto for helpful discussions and insights.

\FloatBarrier

\bibliographystyle{plainnat}
\bibliography{references}

\newpage

\input{supplementary.tex}

\end{document}

%% file: macros.tex
\usepackage{amsmath,amsthm,amssymb}
\usepackage{xspace}

\newcommand\cut[1]{}

\newcolumntype{C}[1]{>{\centering\arraybackslash}m{#1}}
\newcolumntype{R}[1]{>{\raggedleft\arraybackslash}m{#1}}

\newcommand{\myvec}[1]{\mathbf{#1}}
\newcommand{\myvecsym}[1]{\boldsymbol{#1}}

\newcommand{\valpha}{\myvecsym{\alpha}}

\newcommand{\vc}{\myvec{c}}

\newcommand{\vm}{\myvec{m}}
\newcommand{\vmh}{\myvec{{\hat{m}}}}

\newcommand{\vs}{\myvec{s}}

\newcommand{\vx}{\myvec{x}}
\newcommand{\vxh}{\hat{\vx}}

\newcommand{\vz}{\myvec{z}}

\newcommand{\be}{\begin{equation}}
\newcommand{\ee}{\end{equation}}
\newcommand{\bea}{\begin{eqnarray}}
\newcommand{\eea}{\end{eqnarray}}
\newcommand{\beaa}{\begin{eqnarray*}}
\newcommand{\eeaa}{\end{eqnarray*}}

\DeclareMathAlphabet{\mathpzc}{OT1}{pzc}{m}{n}

%% file: introduction.tex
\section{Introduction}\label{S:intro}

Realistic visual scenes contain rich structure, which humans effortlessly exploit to reason effectively and intelligently. In particular, object perception, the ability to perceive and represent individual objects, is considered a fundamental cognitive ability that allows us to understand -- and efficiently interact with -- the world as perceived through our senses \citep{Johnson2018-dg,Green2017-ap}. However, despite recent breakthroughs in computer vision fuelled by advances in deep learning, learning to represent realistic visual scenes in terms of objects remains an open challenge for artificial systems.

The impact and application of robust visual object decomposition would be far-reaching.
Models such as graph-structured networks that rely on hand-crafted object representations have recently achieved remarkable results in a wide range of research areas, including reinforcement learning, physical modeling, and multi-agent control \citep{Battaglia2018-yl, wang_2017, hamrick_2017, hoshen_2017}.
The prospect of acquiring visual object representations through unsupervised learning could be invaluable for extending the generality and applicability of such models.

Most current approaches to object decomposition involve supervision, namely explicitly labeled segmentations in the dataset \citep{ronneberger2015u, jegou2017one, he2017mask}.
This limits the generalization of these models and requires ground-truth segmentations, which are very difficult to acquire for most datasets.
Furthermore, these methods typically only segment an image and don't learn structured object representations.
While some unsupervised methods for scene decomposition have been developed, their performance is limited to very simple visual data \citep{Greff2016tagger, greff2017nem, van2018relational, Eslami2016}. On the other hand, Generative Query Networks \citep{Eslami2018} have demonstrated impressive modelling of rich 3D scenes, but do not explicitly factor representations into objects and are reliant on privileged view information as part of their training.

Recent progress has been made on learning object representations that support feature compositionality, for which a variety of VAE-based methods have become state-of-the-art \citep{higgins2017, Kim_Mnih_2017, Chen_etal_2016, Locatello_etal_2018}. 
However, these methods ignore the structural element of objects, hence are limited to simple scenes with only one predominant object.

We propose that good representations of scenes with multiple objects should fulfill the following desiderata:
\begin{itemize}
    \item A common representation space used for each object in a scene.
    \item Ability to accurately infer objects in 3-dimensional scenes with occlusion.
    \item Flexibility to represent visual datasets with a variable number of objects.
    \item Generalise at test time to (i) scenes with a novel number of objects (ii) objects with novel feature combinations, and (iii) novel co-occurrences of objects.
\end{itemize}

Here, we introduce an architecture that learns to segment and represent components of an image.
This model includes a segmentation network and a variational autoencoder (VAE) \citep{Kingma_Welling_2014, Rezende_etal_2014} trained in tandem. It harnesses the efficiency gain of operating on an image in terms of its constituent objects to decompose visual scenes.

We call this model the \textbf{Multi-Object Network (\ourawesomemodel)} and apply it to a variety of datasets, showing that it satisfies all of our aforementioned desiderata.
Our key contributions are:
\begin{enumerate}
    \item An unsupervised generative model for visual scenes.
    \item State-of-the-art decomposition performance on non-trivial 3D scenes, including generalisation and occlusion-handling.
    \item Ability to learn disentangled representations of scene elements in a common latent code.
\end{enumerate}

%% file: model.tex
\section{Method}\label{S:method}

\subsection{The Multi-Object Network}\label{S:monet}
\begin{figure}[t!]
  \centering
  \includegraphics[width=0.98\linewidth]{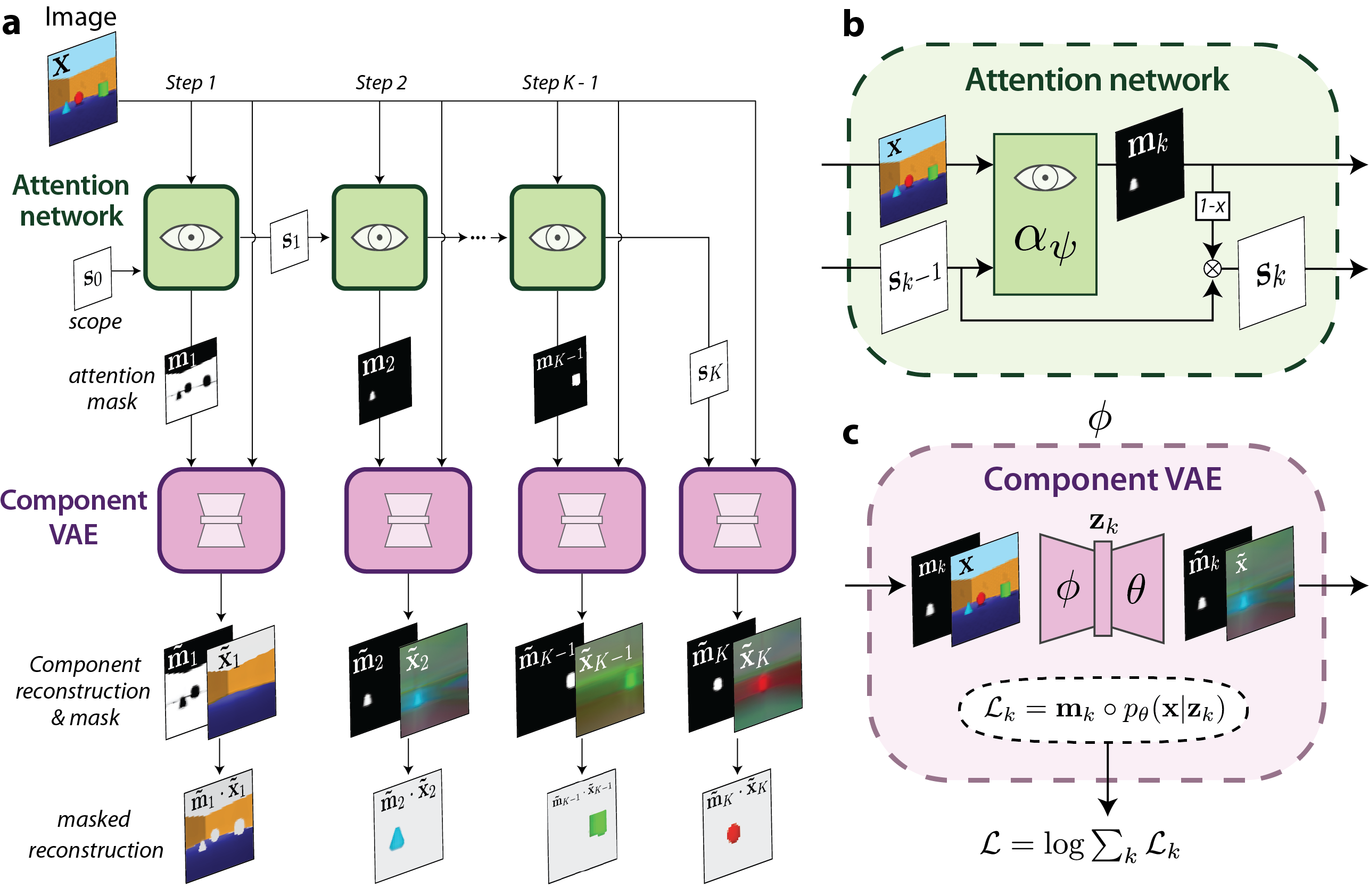}
  \caption{\textbf{Schematic of \ourawesomemodel.} \textbf{(a)} Overall compositional generative model architecture. The attention net recurrently generates the masks over a sequence of steps to condition the component VAE, labelling which pixels to focus on representing and reconstructing for that component. \textbf{(b)} Recursive decomposition process: the attention network at a particular step is conditioned on the image and the current \emph{scope}, which is what currently remains to be explained of the scene (with the initial scope $\vs_0 = \myvecsym{1}$). The attention mask outputted will be some portion of the scope, and the scope for the next step is then what still remains to be explained after accounting for this attention mask. \textbf{(c)}. The component VAE receives both the image and a mask as input, and is pressured only to model the masked region by applying the mask to weight the component likelihood in the loss. Thus the reconstruction component is unconstrained outside of the masked region, enabling it for example to fill in occluded regions. The VAE also models the masks themselves. See main text for more details.
  }
  \label{fig:schematic}
\end{figure}
The Multi-Object network (\ourawesomemodel) models scenes compositionally, by spatially decomposing a scene into parts and modelling each of those parts individually over a set of `slots' with a common representation code (Figure~\ref{fig:schematic}). An attention module provides spatial masks corresponding to all the parts for a given scene, while a component VAE independently models each of the parts indicated by the masks.

The component VAE is a neural network, with an encoder parameterised by $\phi$ and a decoder parameterised by $\theta$ (see Figure~\ref{fig:schematic}c). The encoder parameterises a distribution over the component latents $\vz_k$, conditioned on both the input image $\vx$ and an attention mask $\vm_k$. The mask indicates which regions of the image the VAE should focus on representing via its latent posterior distribution, $q_{\phi}(\vz_k|\vx, \vm_k)$. Crucially, during training, the VAE's decoder likelihood term in the loss $p_\theta(\vx | \vz_k)$ is weighted according to the mask, such that it is unconstrained outside of the masked regions. A complete image is compositionally modelled by conditioning with a complete set of attention masks for the image (i.e. $\sum_{k = 1}^{K} \vm_k = \myvecsym{1}$), over $K$ independent passes through the VAE.

The VAE is additionally required to model the attention masks over the $K$ components, where their distribution $p(\vc | \{\vm_k\})$ is the probability that pixels belong to a particular component $k$, i.e. $\vm_k = p(\vc = k | \{\vm_k\})$. In \ourawesomemodel, the mask distribution is learned by the attention module, a neural network conditioned on $\vx$ and parameterised by $\psi$. Thus, we refer to this distribution as $q_\psi(\vc | \vx)$. The VAE's generative model of those masks is denoted as $p_\theta(\vc | \{\vz_k\})$.

We wanted \ourawesomemodel to be able to model scenes over a variable number of slots, so we used a recurrent attention network $\alpha_\psi$ for the decomposition process. In particular, we arranged the recurrence as an autoregressive process, with an ongoing state that tracks which parts of the image have yet to be explained (see Figure~\ref{fig:schematic}b). We call this state the scope $\vs_k$, which is an additional spatial mask updated after each attention step. Specifically, it signifies the proportion of each pixel that remains to be explained given all previous attention masks, where the scope for the next step is given by:

\begin{equation}
    \mathbf{s}_{k + 1} = \vs_k \big(1 - \alpha_\psi(\vx; \vs_k)\big)
\end{equation}
with the first scope $s_0 = \myvecsym{1}$. The attention mask for step $k$ is given by:

\begin{equation}
    \vm_k = \vs_{k - 1}\alpha_\psi(\vx; \vs_{k - 1})
\end{equation}
except for the last step $K$, where the attention network is not applied, but the last scope is used directly instead, i.e. $\vm_K = \vs_{K - 1}$. This ensures that the entire image is explained, i.e. $\sum_{k = 1}^{K} \vm_k = \myvecsym{1}$.

The whole system is trained end-to-end with a loss given by:
\begin{align}\label{monet_loss}
\begin{autobreak}
    \mathcal{L}(\phi; \theta; \psi; \vx) =
    - \log\sum_{k = 1}^{K} \vm_k p_\theta(\vx | \vz_k)
    + \beta \kldiv{ \prod_{k=1}^{K} q_{\phi}(\vz_k|\vx, \vm_k)}{p(\vz)}
    + \gamma \kldiv{q_\psi(\vc | \vx)}{p_\theta(\vc | \{\vz_k\})}
\end{autobreak}
\end{align}

The first two terms of the loss are derived from the standard VAE loss. The first term is the decoder negative log likelihood, given our mixture of components decoder distribution, as discussed above. The second term is the Kullback–Leibler divergence (KL) divergence of the latent posterior (factorised across slots) with the latent prior, weighted with a hyperparameter $\beta$, following \cite{higgins2017}, which can be tuned to encourage learning of disentangled latent representations. The last term to be minimised is the KL divergence between the attention mask distribution $q_\psi(\vc | \vx)$ and the VAE's decoded mask distribution $p_\theta(\vc | \{\vz_k\})$. This is also weighted by a tuneable hyperparameter  $\gamma$ that here modulates how closely the VAE must model the attention mask distribution.

\subsection{Exploiting compositional structure}\label{S:exploiting_composition}
In this section we aim to motivate the development of the approach described above, specifically exploring the reasons we might expect the loss defined in Eq.~\ref{monet_loss} to decrease if masks corresponding to semantically meaningful decompositions are learned. This includes an empirical confirmation, for which we construct a visually interesting 3D scenes dataset (also used in our main results) and compare performance under different decomposition regimes. However, the main results for unsupervised scene decomposition with \ourawesomemodel are in the next section (Section.~\ref{S:results}).

We originally developed this architecture while considering how to learn to meaningfully decompose a scene without supervision. We wanted to identify some general consequences of the compositional structure of scenes that could push the optimisation towards decomposition. We started from the hypothesis that compositional visual scenes can be more efficiently processed by something like a deep neural network if there is some common repeating structure in scenes that can be exploited. In particular, if a network performing some task can be repeatedly reused across scene elements with common structure (such as objects and other visual entities), its available capacity (limited for example by its architecture and weights) will be more effectively utilised and thus will be more efficient than the same network processing the entire scene at once. This both motivates a key benefit of decomposition -- identifying ways to break up the world that can make a range of tasks more efficient to solve -- \emph{and} suggests an approach for identifying the compositional structure of data.

\begin{figure}[t!]
  \centering
  \includegraphics[width=1.0\linewidth]{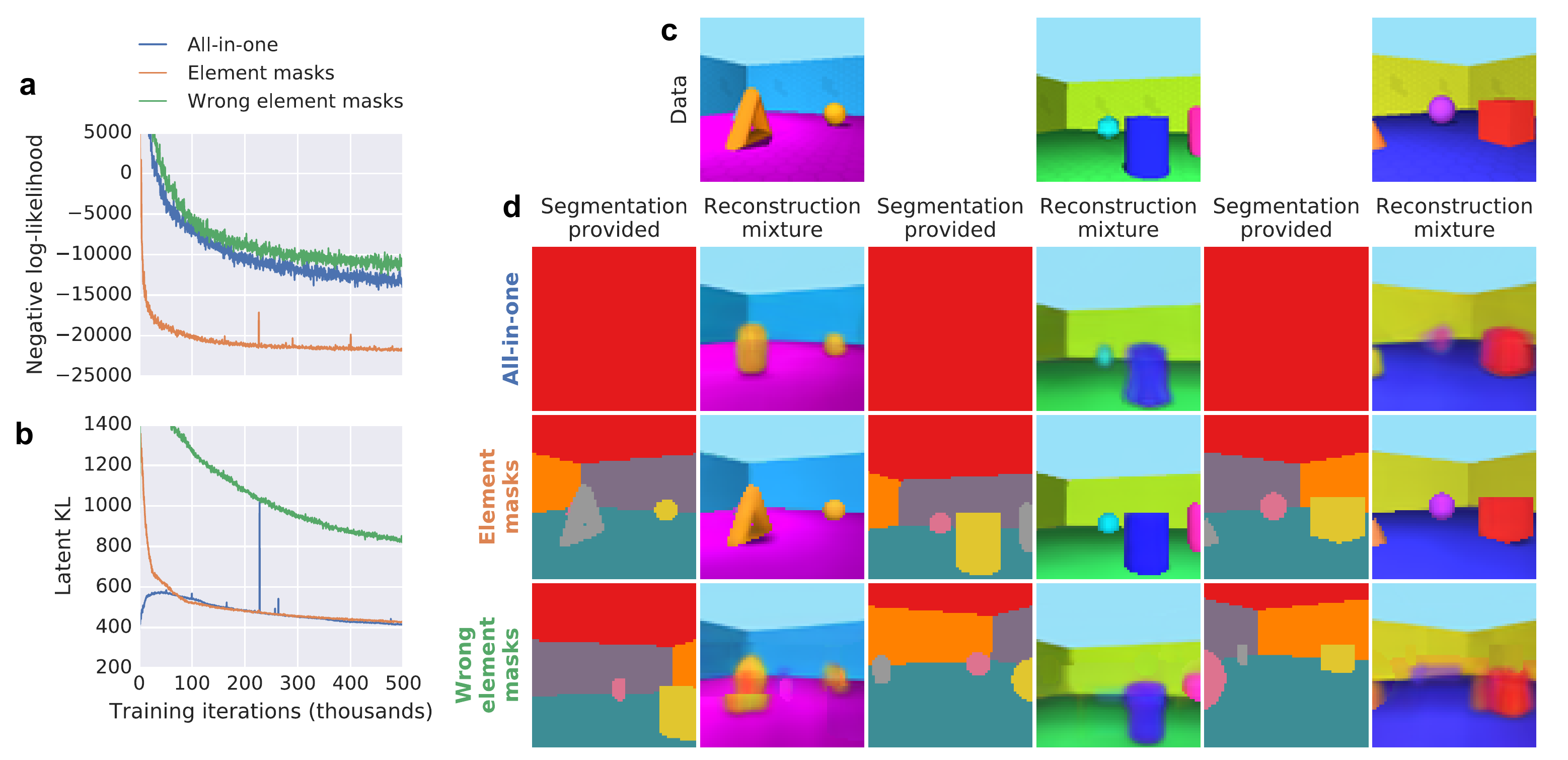}
  \caption{\textbf{Semantic decomposition improves reconstruction accuracy.} These are results from experiments to motivate the \ourawesomemodel training objective by replacing the learned masks from the attention network with provided masks (see Section.~\ref{S:exploiting_composition}). The component VAE is trained to reconstruct regions of \oir images given the provided masks. Three mask conditions are compared: reconstructing everything in the first pass (\allinone), reconstructing individual scene elements in separate passes using ground-truth masks (\elementmasks), and finally a control condition requiring reconstruction of regions given by element masks for the wrong scene (\wrongelementmasks). \textbf{(a)} and \textbf{(b)} show the negative log-likelihood (mask-weighted sum over scene elements) and KL divergence (with the latent prior), respectively over training for the three mask conditions. \textbf{(d)} Example masks for each of the conditions (summarised as colour-coded segmentation maps), for example scenes shown in \textbf{c}, with corresponding reconstructions after training under the different conditions (reconstructions rendered by mixing the scene reconstruction components according to the masks). The colors used in the segmentation visualization are independent of the colors in the scenes themselves.}
  \label{fig:oir_decomposition}
\end{figure}

In the context of our scene representation learning goal, this hypothesis would predict that a network tasked with autoencoding visual scenes would perform better, \emph{if} it is able to build up scenes compositionally by operating at the level of the structurally similar scene elements. Specifically, such a network should have a lower reconstruction error than if the same network were instead trained to reconstruct entire scenes in a single pass. 

\paragraph{Empirical validation.}
To test this hypothesis, we constructed a dataset of rendered 3D scene images of non-trivial visual complexity, the \oir dataset (see Figure~\ref{fig:objects_in_room_dataset} for example images). We specifically designed the scenes to be composed of multiple scene elements that share varying degrees of common structure. The \oir images were generated as randomised views inside a cubic room with three randomly shaped, coloured and sized objects scattered around the room. For any given image, 1-3 of those objects are in view. The wall and floor colour of the room in each image are also randomised (see Section.~\ref{S:datasets} in the appendix for more details). For each image $\vx$, a set of 7 ground-truth spatial masks $\vmh$ was also generated, indicating the pixel extents of visual elements comprising the scene (the floor, sky, each of two adjoining wall segments, and three objects).

We tested the hypothesis by training the model outlined above on this dataset, but instead of learning the masks as in \ourawesomemodel we provided a set of seven attention masks $\{\vm_k\}$ masks under three different conditions of interest. In the \allinone condition, the entire image was always segmented into the first mask (i.e. $\vm_1 = \myvecsym{1}$ with the remaining masks being all zeros). In the \elementmasks condition, the scene was segmented according to ground-truth visual element masks, i.e. $\{\vm_k\} = \{\vmh_k\}$. Finally, the \wrongelementmasks is a control condition with the mask in the same format as \elementmasks, but always for a different, random scene (i.e. always an incorrect segmentation, but with the same statistics as \elementmasks).

Figure~\ref{fig:oir_decomposition} shows the results of training the component VAE under each condition. As predicted, the reconstructions for the model trained with \elementmasks are significantly better than the reconstructions from the model trained with \allinone masks. This can be seen by the large gap between their reconstruction errors over the course of training (see the negative log-likelihoods in Figure~\ref{fig:oir_decomposition}a). The \elementmasks reconstructed images are also markedly better visually (see the mixtures of the reconstruction components in Figure~\ref{fig:oir_decomposition}d), with objects from \allinone in particular being more blurred, and sometimes missing (e.g. in the middle example in Figure~\ref{fig:oir_decomposition}d, the \allinone model fails to reconstruct the magenta object). In contrast, the latent KLs for the two conditions were very similar (see Figure~\ref{fig:oir_decomposition}b).

However, the kind of segmentation is important: the model trained with \wrongelementmasks (where the masks do not correspond to structurally meaningful visual elements for the relevant scene) performs worse than both of the other conditions. Reconstruction error and latent KL are significantly higher (gold curves in Figure~\ref{fig:oir_decomposition}a and Figure~\ref{fig:oir_decomposition}b), and the reconstructed images have clear edge artifacts where the provided masks are misaligned with the structural elements in the scene (bottom row in Figure~\ref{fig:oir_decomposition}d).

Overall, these results support the hypothesis that processing elements of scenes in a way that can exploit any common structure of the data makes more efficient use of a neural network's capacity. Furthermore, the significant improvement seen in the loss suggests this corollary can be used to discover such structurally aligned decompositions without supervision. In the next section, we show the results from the full \ourawesomemodel setup, in which the attention masks are learned in tandem with the object representations, in a fully unsupervised manner.
% In the appendix section Section ~\ref{S:decomposition_factors} we further explore some of the possible factors that could be contributing to the efficiency improvement, and show further empirical results that support more efficient use of limited network capacity as the most important factor behind the gain. 

%% file: results.tex
\section{Results}\label{S:results}
For all experiments we used the same basic architecture for \ourawesomemodel. The attention network used an architecture inspired by the U-Net \citep{ronneberger2015u}. For the VAE, we used an encoder with convolutional followed by fully connected layers to parameterise a diagonal Gaussian latent posterior (with a unit Gaussian prior), together with a spatial broadcast decoder \citep{watters2019} to encourage the VAE to learn disentangled features. The decoder parameterised the means of pixel-wise independent Gaussian distributions, with fixed scales. All results are shown after training for 1,000,000 iterations. See Section.~\ref{S:hypers} in the appendix for more details on the architecture and hyperparameters used.

\begin{figure}[t!]
  \centering
  \includegraphics[width=0.8\linewidth]{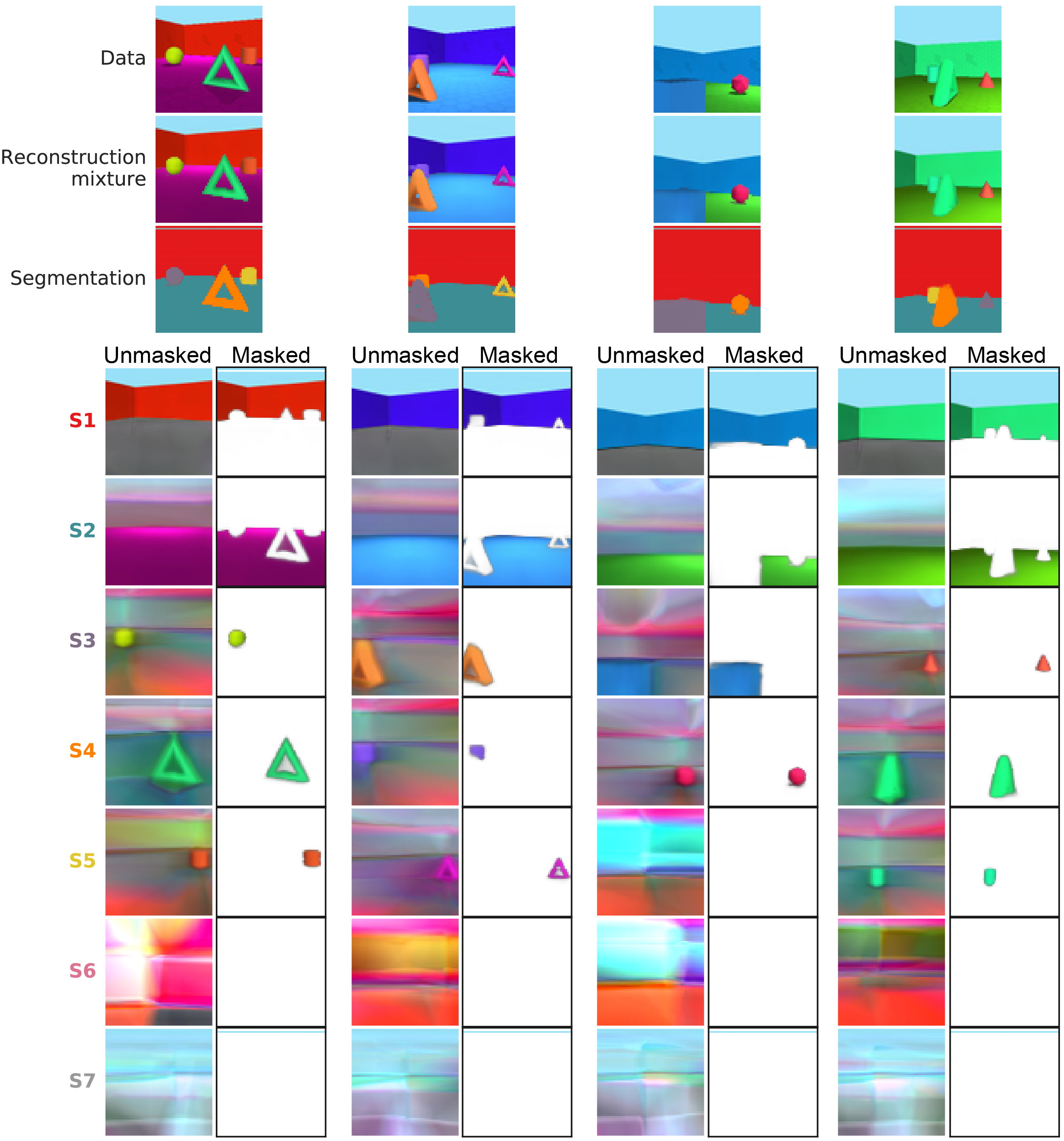}
  \caption{\textbf{Unsupervised decomposition results on the \oir dataset.} \newline Results from \ourawesomemodel trained on a dataset of 3D scenes of a room with 1-3 objects in view. Each example shows the image fed as input data to the model, with corresponding outputs from the model. Reconstruction mixtures show sum of components from all slots, weighted by the learned masks from the attention network. Colour-coded segmentation maps summarising the attention masks $\{\vm_k\}$ are shown in the next row. Rows labelled S1-7 show the reconstruction components of each slot. Unmasked versions are shown side-by-side with corresponding versions that are masked with the VAE's reconstructed masks $\tilde{\vm}_k$. In the third example, note the correct handling of the large blue cube with the same colour as the background. }
  \label{fig:monet_oir}
\end{figure}

\begin{figure}[t!]
  \centering
  \includegraphics[width=0.9\linewidth]{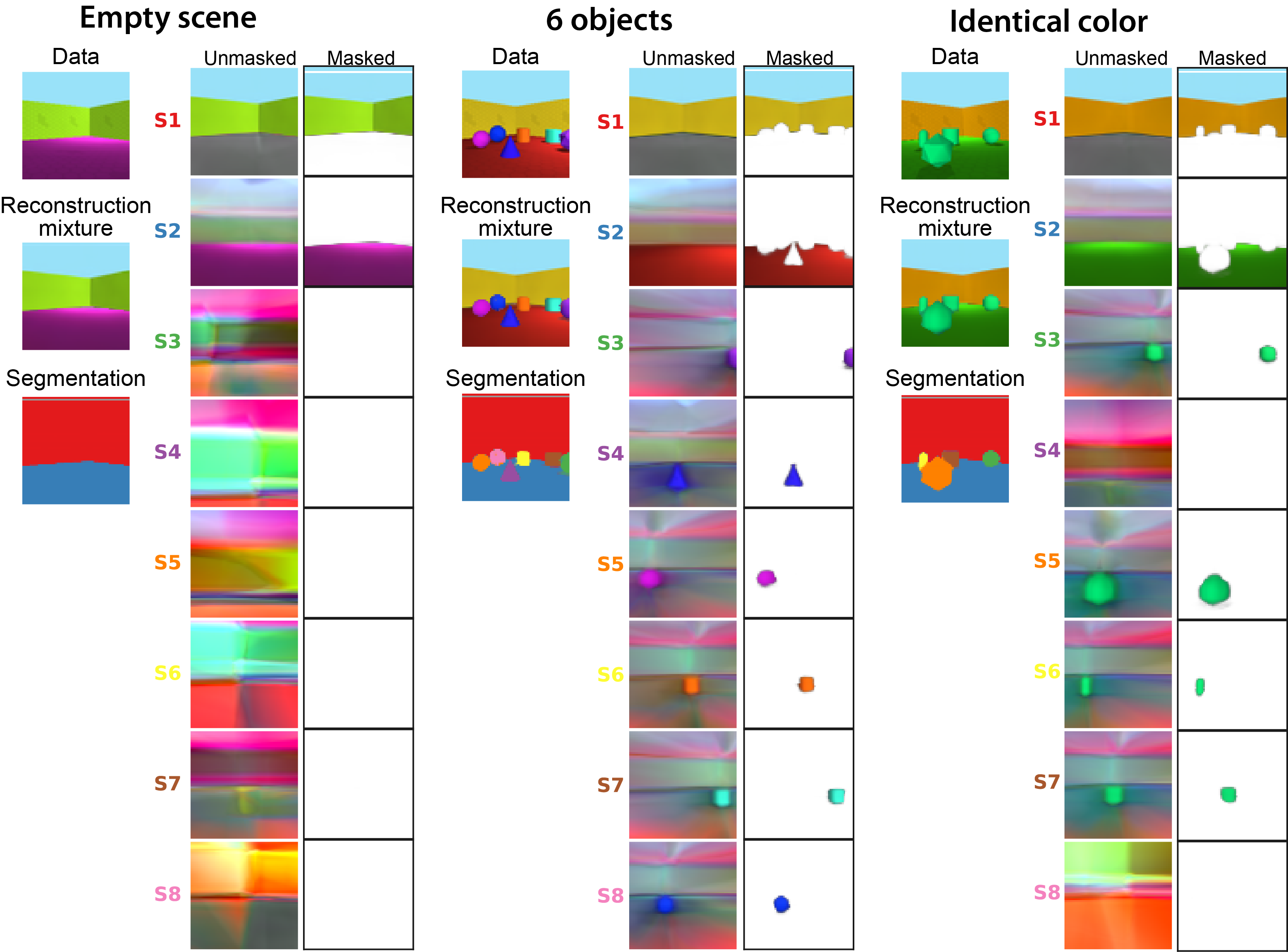}
  \caption{\textbf{\ourawesomemodel generalisation outside of training regime and data distribution.} \newline Results presented in a similar format and using the same trained model as as Figure~\ref{fig:monet_oir} but now run for 9 slots (of which only slots 1-8 are shown). The model robustly segments novel scenes with no objects (left), or with 6 objects (middle) -- double the number of objects seen at training time (utilising extra test time slots). It also correctly handles scenes with 4 identically coloured and visually overlapping objects (right). }
  \label{fig:monet_oir_generalisation}
\end{figure}

\subsection{Results on \oir}
We trained \ourawesomemodel with $K = 7$ slots on the \oir dataset with 1-3 objects in view (introduced in Section.~\ref{S:exploiting_composition}); results are shown in Figure~\ref{fig:monet_oir}. \ourawesomemodel learns to generate distinct attention masks for the floor, wall + sky (combined), and the individual objects (including some of their shadows), across a variety of scenes as can be seen in the segmentation maps (\emph{Segmentation} row in Figure~\ref{fig:monet_oir}a).

The unmasked reconstruction components of each slot are shown side-by-side with masked versions (see rows labelled S1-S7 Figure~\ref{fig:monet_oir} show the respective unmasked reconstruction components). The latter are masked using the VAEs mask reconstructions $\tilde{\vm}_k$, and show accurate reconstructions within those regions as well as demonstrating the VAEs modelling of the attention mask distribution. The unmasked components reveal coherent filling in over regions otherwise occluded by objects (e.g. floors and walls). Where slots have empty masks, their reconstructions components tend to contain unspecific background patterns. Combining reconstruction components according to their attention masks yields high quality reconstructions of the full scenes (\emph{Reconstruction mixtures} in Figure~\ref{fig:monet_oir}). 

We also assessed the generalisation capabilities of \ourawesomemodel, by testing with extra decomposition steps on novel scenes (Figure~\ref{fig:monet_oir_generalisation}). Using the recurrent autoregressive process of decomposition, we were able to run the same trained model but for more attention steps at test time, here extending it to a 9 slot model (although only the first 8 are shown, with more extensive results shown in supplementary Figure~\ref{fig:monet_oir_supp}). This model generalised well to scene configurations not observed during training. We presented empty room scenes, scenes with 6 objects, and scenes with 4 objects of the same colour on a similarly coloured floor. In each case, the model produced accurate segmentations and filled up extra test time slots with the excess objects as appropriate.

\begin{figure}[t!]
  \centering
  \includegraphics[width=1.0\linewidth]{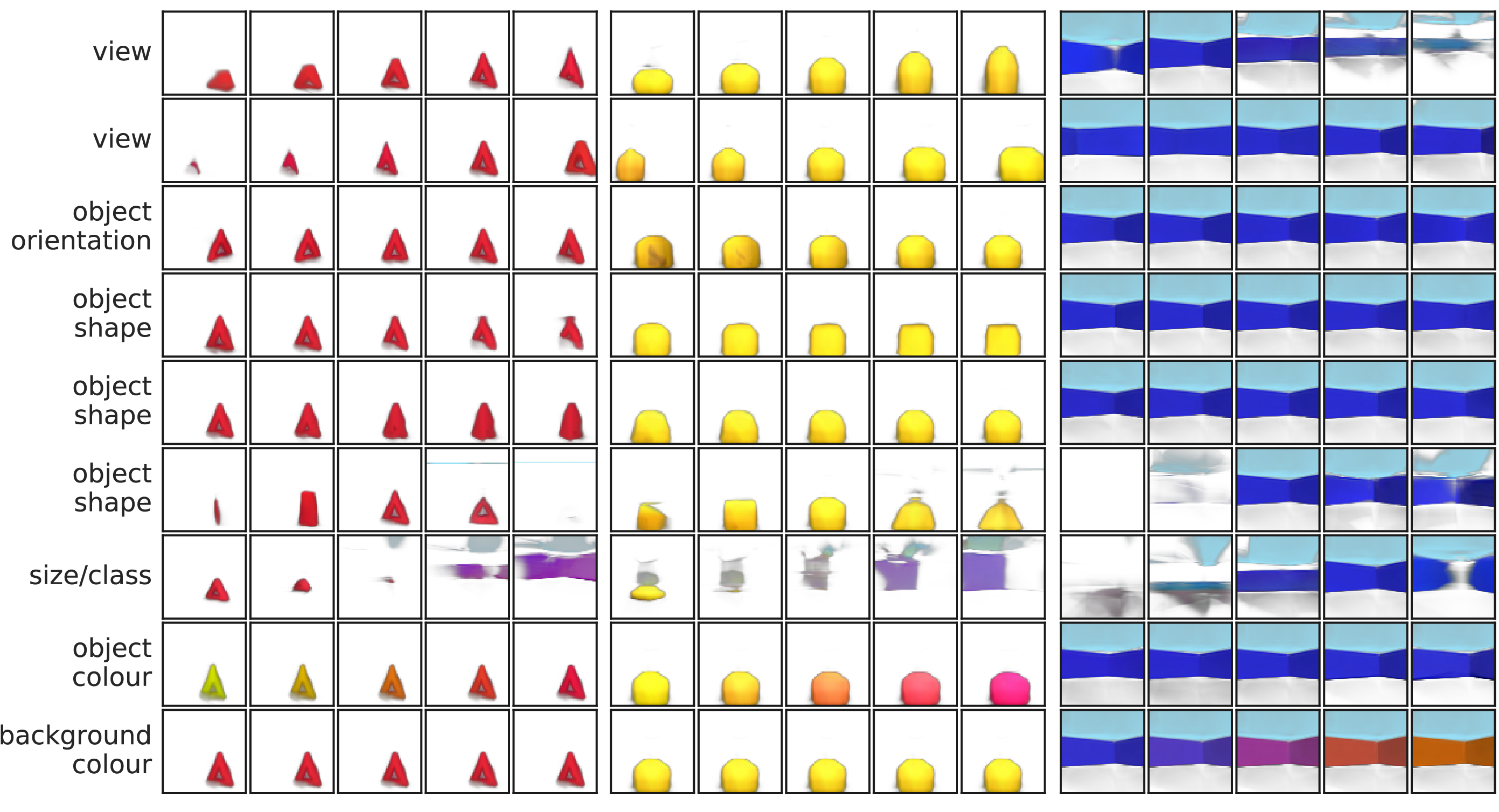}
  \caption{\textbf{\ourawesomemodel object representations learned on \oir.} Each plot (left, middle, and right) shows how the reconstruction component of a particular slot (with its reconstructed mask applied) changes as each single latent $z_{k,i}$ is traversed from -1.0 to +1.0. Same trained model as in Figure~\ref{fig:monet_oir}. Results from different seed images and slots shown in each column group. The left and middle column groups show components from slots representing scene objects, and the right column shows a wall + sky component. A subset of latents were picked based on visual inspection and are consistent across column groups. Labels show their intuitive feature interpretations. Note here we used reconstructed masks that were not normalised across other components to generate the components in isolation. }
  \label{fig:oir_disentangling}
\end{figure}

\subsection{Disentangled representations}
An important design feature of \ourawesomemodel is that it learns a latent representation unified across slots. As discussed above, we utilised a weight modulating the latent KL in the loss \citep{higgins2017, burgess2018} and a broadcast decoder \citep{watters2019} in \ourawesomemodel to encourage disentangling of the latent representation at the feature level.

We assessed the feature-level disentangling of the latent representations $\vz_k$ by seeding input images and inspecting how the reconstruction components changed as single latents $\vx_{k,i}$ were traversed (Figure~\ref{fig:oir_disentangling}). These results are generated from the same trained model as above. We show the components with their reconstructed masks applied to incorporate how the latent traversals also affect them (note in this figure we use masks not normalised by other components). From visual inspection, we identified many latents independently controlling interpretable features of each visual element, indicative of disentangling. Some latents only controlled specific features in slots containing scene objects (those prefixed with `object', e.g. the shape latents), or were specific to slots containing the wall + sky component (e.g. the distinct `background' and `object' colour latents). Others controlled features across all visual element classes (e.g. the `view' latents), or switched between them (the `size/class' labelled latent).

\begin{figure}[t!]
  \centering
  \includegraphics[width=1.0\linewidth]{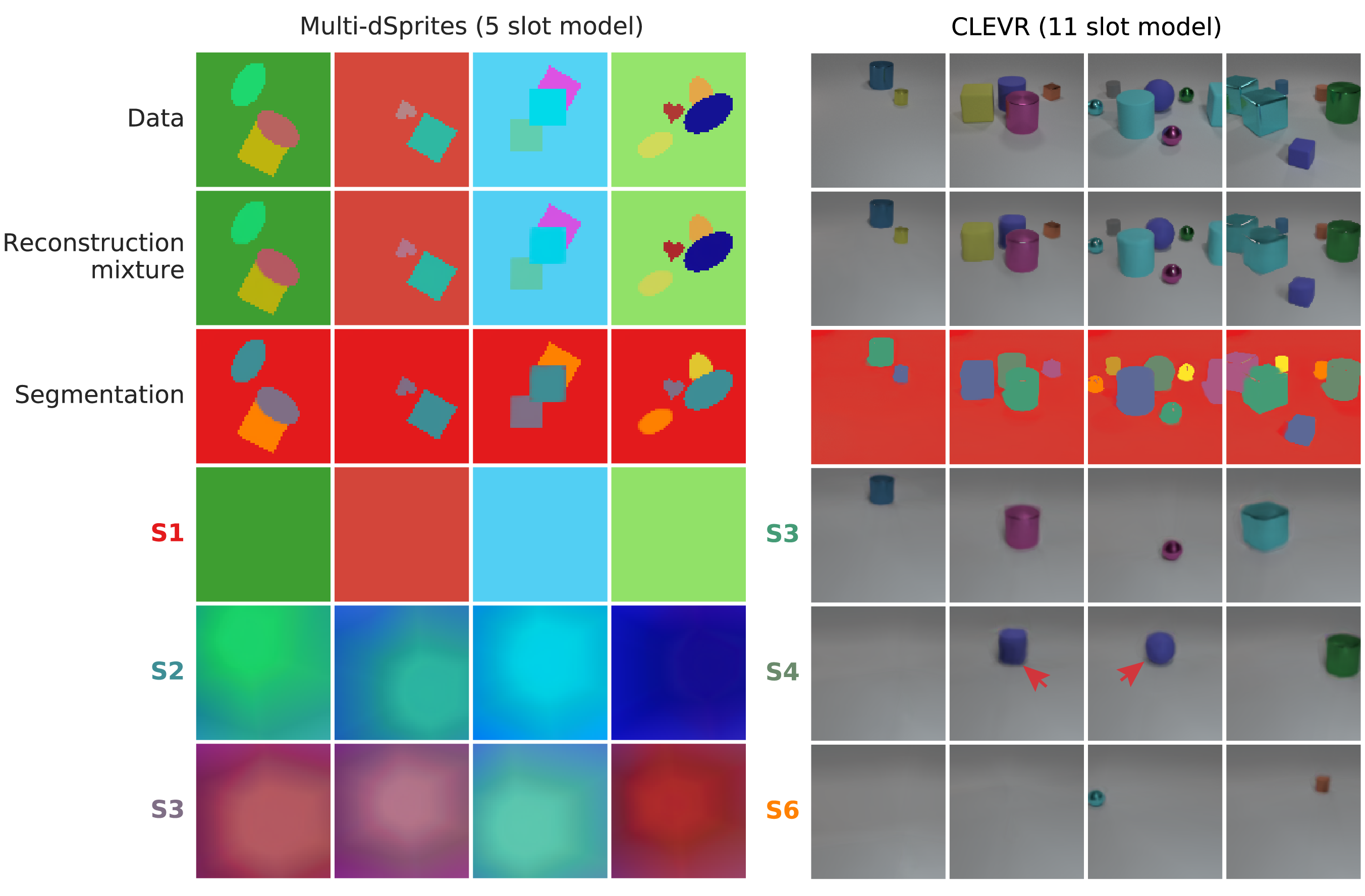}
  \caption{\textbf{\ourawesomemodel decomposition on \multidsprites and \clevr images.} Format as in Figure~\ref{fig:monet_oir}. \textbf{Left panel} shows results from model with five slots trained on \multidsprites (with 1-4 sprites per image). Unmasked component reconstructions from first three slots are shown. Note the correct segmentation. \textbf{Right panel} shows results from 11 slot model trained on images from the \clevr dataset \citep{Johnson2017clevr}. Unmasked component reconstructions from three representative slots are shown. Red arrows highlight occluded regions of shapes that are completed as full shapes. Rightmost column demonstrates accurate delineation of two very similar visually overlapping shapes.}
  \label{fig:monet_other_datasets}
\end{figure}
\subsection{Results on \multidsprites}
\ourawesomemodel was also able to accurately decompose 2D sprite scenes into the constituent sprites (see the left panel Figure~\ref{fig:monet_other_datasets}). To create this dataset, we adapted the dSprites \citep{dsprites17} dataset by colourising 1-4 randomly chosen dSprites and compositing onto a single image with a uniform randomly coloured background (with occlusions; see  Section.~\ref{S:datasets} in the appendix for more details). The same model architecture and hyperparameters as before were used (except with $K = 5$ slots), enabling us to verify robustness across very different datasets. After training, the model was able to distinguish individual sprites and the background robustly, even those sprites behind multiple occlusions or very difficult to distinguish by eye. As individual unmasked reconstruction components generally consisted of the entire image taking the relevant sprite's colour, we do not show the masked versions for brevity. More examples can be seen in supplementary Figure~\ref{fig:monet_dsprites_supp}
\subsection{Results on \clevr}
We tested \ourawesomemodel on images from \clevr \citep{Johnson2017clevr}, a dataset of simple 3D rendered objects (see right panel in Figure~\ref{fig:monet_other_datasets}). We down-sampled the images to 128x128, and cropped them slightly to ensure more of the frame was occupied by objects (see Section.~\ref{S:datasets} in the appendix for more details), resulting in between 2-10 visible objects per image. To handle the increased resolution of this dataset, we added extra blocks to the attention network (see Section.~\ref{S:hypers} in the appendix for more details) and used $K = 11$ slots, but left all other hyperparameters unchanged.

The model robustly segmented scenes across the diversity of configurations into single objects and generated high quality reconstructions (right panel in Figure~\ref{fig:monet_other_datasets}; and also see more extensive examples in supplementary Figure~\ref{fig:monet_clevr_supp}).

The model is also robust to the relatively frequent occlusions between objects in this dataset. Interestingly, the unmasked reconstruction components of occluded objects showed very coherent filling in of occluded regions, demonstrating how \ourawesomemodel is learning from and constrained by the structure of the data (red arrows in Figure~\ref{fig:monet_other_datasets} highlights some examples). The model was also able to correctly distinguish overlapping shapes with very similar appearances (see rightmost column in Figure~\ref{fig:monet_other_datasets}).

%% file: related_work.tex
\section{Related Work}\label{S:related}
\subsection{Supervised Approaches} 
Semantic segmentation using supervision from pixel-wise class labels is a well-studied problem. Neural architectures such as the U-Net \citep{ronneberger2015u}, Fully Convolutional DenseNet \citep{jegou2017one}, and DeepLabv3 \citep{chen2018deeplab}) have successively advanced the state of the art on this domain, furnishing strong inductive biases for image segmentation. \ourawesomemodel can utilize these architectures in place of its attention module (to propose viable object masks) even without a supervised training signal.

While \ourawesomemodel is partly about instance segmentation, its recurrent decomposition process bears little similarity to state-of-the-art instance segmentation models like Mask R-CNN \citep{he2017mask} and PANet \citep{liu2018path}. These are trained via supervision to propose bounding boxes for all objects in a scene at once. Separate modules then process the bounding boxes one at a time to yield object masks and classify the objects. The overarching distinction with our approach is that features learnt by instance segmentation methods only help identify object classes, not explain away the complexity of the scene. \ourawesomemodel also benefits from learning segmentations and representations jointly, which allows it to discover fine-grained object masks directly and without supervision. Admittedly, Mask R-CNN and PANet can work with larger, natural images with $\sim$100 objects.

\subsection{Unsupervised Approaches}
A principled approach to deconstruct scenes into underlying components is encapsulated by the vision-as-inverse-graphics paradigm. Methods in this camp \citep{tian2018learning, yao20183d} make domain-specific assumptions about the latent code \citep{NIPS2015_5851} or generative process \citep{wu2017neural} to keep inference tractable, but these assumptions tend to be too strong for the methods to be broadly useful. \ourawesomemodel demonstrates more general scene understanding by learning object features in an unstructured Gaussian latent space via end-to-end learning.

Methods apart from probabilistic and variational inference have also made some headway toward scene understanding. Adversarially trained generative models can demonstrate an implicit understanding of missing parts of a scene by inpainting over them \citep{pathak2016context}. "Self-supervised" models can also show impressive results on object discovery \citep{doersch2015unsupervised} or tracking \citep{Vondrick_2018_ECCV}). But rather than learning from the structure and statistics of the data, these methods often rely on heuristics (such as region masks, neighboring patches, or reference frames for colorization), and occasionally on explicit supervision (e.g. to recognize the presence of a type of object in \cite{NIPS2018_7997}). They largely abstract away the problem of representation learning and have not yielded general principles for representing objects.

That leaves us with a small class of models to which \ourawesomemodel is immediately comparable. The Attend-Infer-Repeat framework \citep{Eslami2016} is the closest to our work in spirit. Like \ourawesomemodel, AIR highlights the representational power of object-wise inference, and uses a recurrent network to attend to one object at a time. Unlike \ourawesomemodel, it explicitly factors an object representation into ‘what’, ‘where’, and ‘presence’ variables. The ‘what’ is modelled by a standard VAE. A spatial transformer module \citep{jaderberg2015spatial} then scales/shifts the generated component to match the ‘where’. Finally, the components are added together (if ‘present’) to form the final image. This additive interaction is restrictive, and as a consequence, neither AIR nor its successor SQAIR \citep{kosiorek2018sequential} can model occluded objects or background pixels. These models have not been shown to scale to larger number of objects.

Another line of work spanning Tagger \citep{Greff2016tagger}, Neural Expectation Maximization \citep{greff2017nem}, and Relational Neural Expectation Maximization \citep{van2018relational} makes use of iterative refinement to decompose scenes into groups of pixels and assign them to individual components. These works draw their intuitive appeal from classical clustering and EM algorithms. But they have not been shown to work beyond small, binarized images or videos. We anticipate \ourawesomemodel can be readily extended to incorporate some form of iterative refinement across components in future work.

% \rish{Other potential related work: probabilistic U-NET. Relation networks, which presuppose objects. DRAW.}

% \todo{Talk about instance segmentation approaches. Talk about segmantic segmentation and how they relate. SOTA instance segmentation models: Mask R-CNN and PANet}

%% file: conclusion.tex
\section{Conclusions and future work}\label{S:conclusion}
We presented \ourawesomemodel, a compositional generative model for unsupervised scene decomposition and representation learning. Our model can learn to decompose a scene without supervision, and learns to represent individual components in a common disentangled code. The approach we take is motivated by a gain in efficiency from autoencoding scenes compositionally, where they consist of simpler coherent parts with some common structure. To the best of our knowledge, \ourawesomemodel is the first deep generative model that can perform meaningful unsupervised decomposition on non-trivial 3D scenes with a varying number of objects such as CLEVR.

We found our model can learn to generate masks for the parts of a scene that a VAE prefers to reconstruct in distinct components, and that these parts correspond to semantically meaningful parts of scene images (such as walls, floors and individual objects). Furthermore, \ourawesomemodel learned disentangled representations, with latents corresponding to distinct interpretable features, of the scene elements. Interestingly, \ourawesomemodel also learned to complete partially occluded scene elements in the VAE's reconstructions, in a purely unsupervised fashion, as this provided a more parsimonious representation of the data. This confirms that our approach is able to distill the structure inherent in the dataset it is trained on. Furthermore, we found that \ourawesomemodel trained on a dataset with 1-3 objects could readily be used with fewer or more objects, or novel scene configurations, and still behave accurately.

This robustness to the complexity of a scene, at least in the number of objects it can contain, is very promising when thinking about leveraging the representations for downstream tasks and reinforcement learning agents.

Although we believe this work provides an exciting step for unsupervised scene decomposition, there is still much work to be done. For example, we have not tackled datasets with increased visual complexity, such as natural images or large images with many objects. As the entire scene has to be `explained' and decomposed by \ourawesomemodel, such complexity may pose a challenge and warrant further model development. It would be interesting, in that vein, to support \emph{partial} decomposition of a scene, e.g. only represent some but not all objects, or perhaps with different precision of reconstruction.

We also have not explored decomposition of videos, though we expect temporal continuity to amplify the benefits of decomposition in principle, which may make \ourawesomemodel more robust on video data.
Finally, although we showed promising results for the disentangling properties of our representations in Figure~\ref{fig:oir_disentangling}, more work should be done to fully assess it, especially in the context of reusing latent dimensions between semantically different components.

%% file: supplementary.tex
\begin{appendices}\label{S:supp}

% \section{Supplementary Materials}

\appendix
\section{Supplementary figures}

\begin{figure}[h]
  \centering
  \vspace{20pt}
  \includegraphics[width=1.0\linewidth]{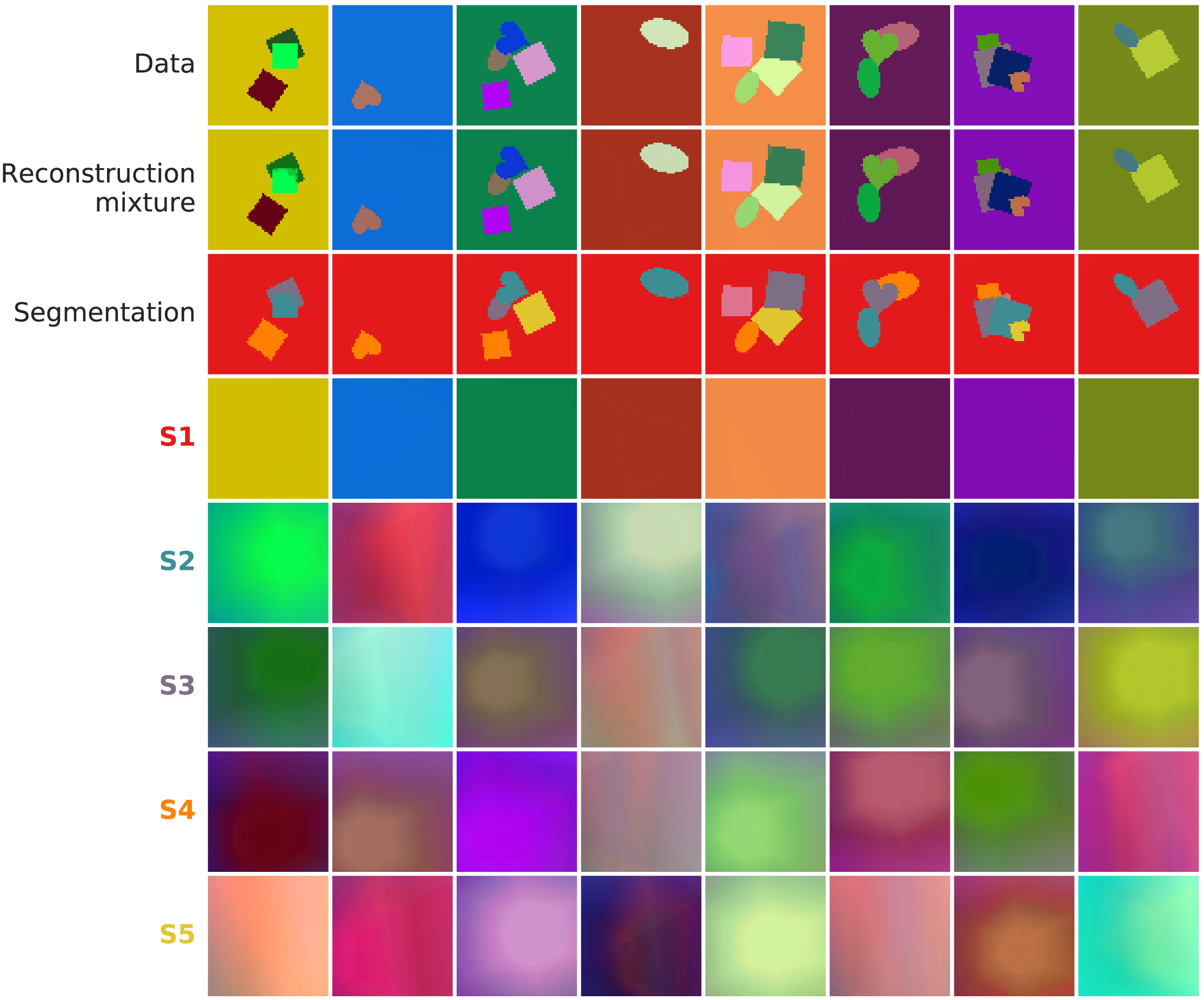}
  \caption{\textbf{Unsupervised \multidsprites decomposition with \ourawesomemodel.} Same trained model and format as Figure~\ref{fig:monet_other_datasets} but showing all 5 slots on randomly selected data samples.}
  \label{fig:monet_dsprites_supp}
\end{figure}

\begin{figure}[t!]
  \centering
  \includegraphics[width=1.0\linewidth]{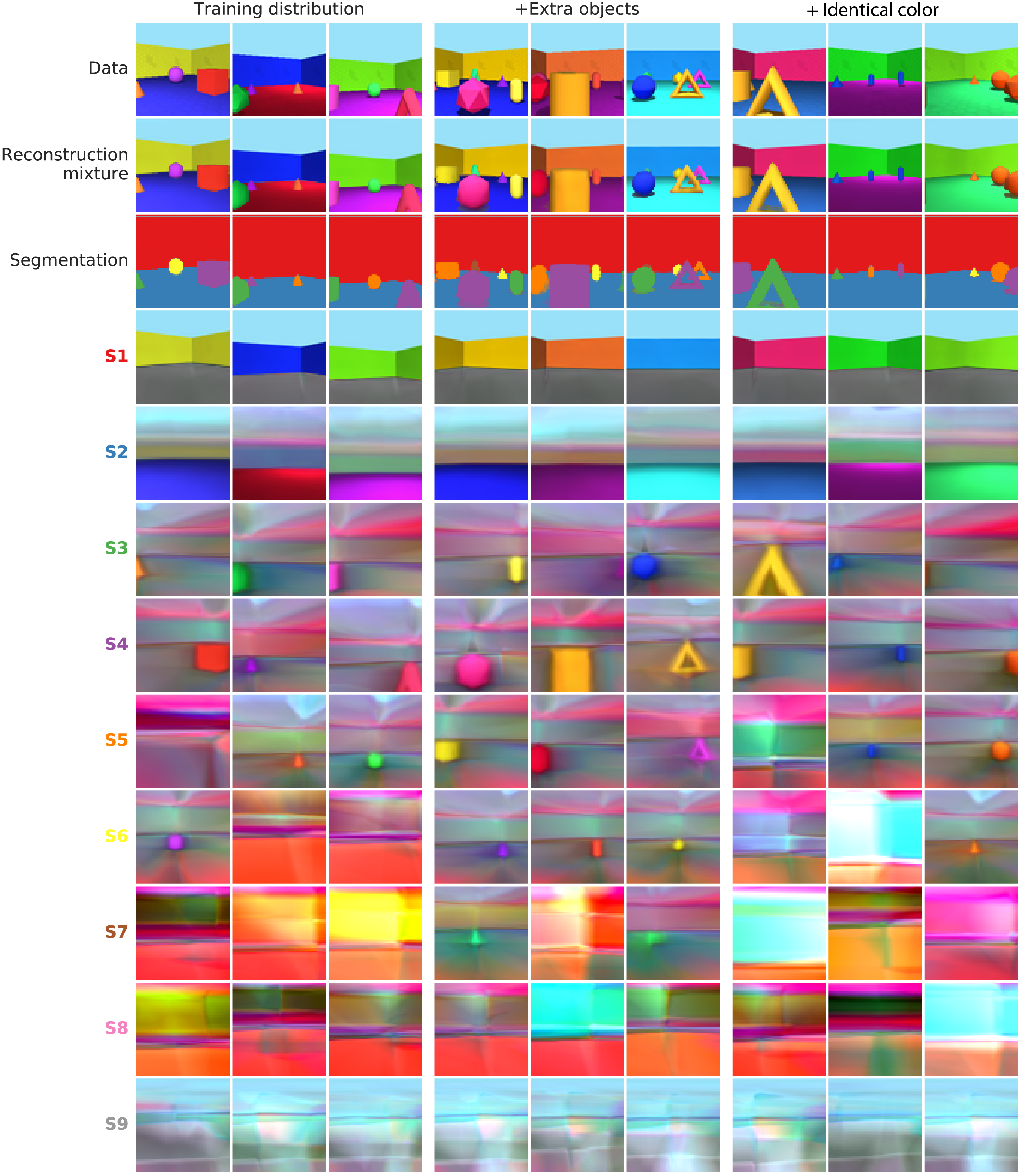}
%   \vspace{-10pt}
  \caption{\textbf{Unsupervised \oir  decomposition with \ourawesomemodel.} Same trained model and format as Figure~\ref{fig:monet_oir} but showing all 9 slots (model trained on 7 slots) on randomly selected data samples. Left three columns show results on the training distribution. Remaining columns shown generalisation results outside of the training distribution: middle three columns on scenes with extra objects; last three columns on scenes with extra objects and all objects the same colour.}
  \label{fig:monet_oir_supp}
\end{figure}

\begin{figure}[t!]
  \centering
  \includegraphics[width=0.95\linewidth]{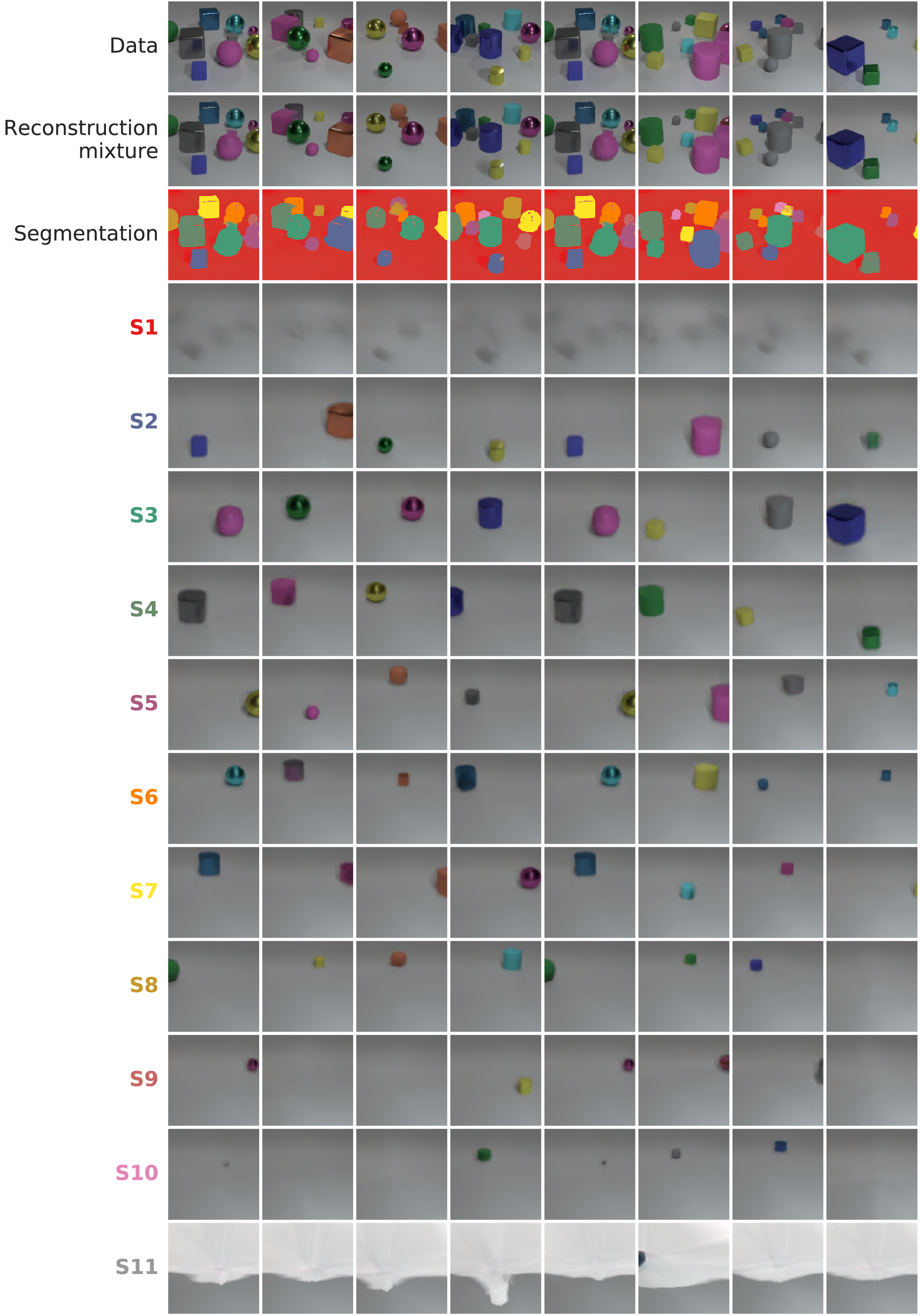}
  \caption{\textbf{Unsupervised \clevr decomposition with \ourawesomemodel.} Same trained model and format as Figure~\ref{fig:monet_other_datasets} but showing all 11 slots on randomly selected data samples.}
  \label{fig:monet_clevr_supp}
\end{figure}

See Figures~\ref{fig:monet_dsprites_supp}--\ref{fig:monet_clevr_supp} for additional examples of \ourawesomemodel decomposition results on the different datasets we considered.

\FloatBarrier
\clearpage

\section{Architecture and hyperparameter details}\label{S:hypers}
\subsection{Component VAE}
The VAE encoder is a standard CNN with 3x3 kernels, stride 2, and ReLU activations. It receives the concatenation of the input image $\vx$ and the attention mask in logarithmic units, $\log \vm_k$ as input. The CNN layers output (32, 32, 64, 64) channels respectively. The CNN output is flattened and fed to a 2 layer MLP with output sizes of (256, 32). The MLP output parameterises the $\myvec{\mu}$ and $\log \mathbf{\sigma}$ of a 16-dim Gaussian latent posterior.

The VAE uses a broadcast decoder \citep{watters2019} to transform the sampled latent vector $\vz_k$ into the reconstructed image component and mask distributions. The input to the broadcast decoder is a spatial tiling of $\vz_k$ concatenated with a pair of coordinate channels -- one for each spatial dimension -- ranging from -1 to 1. These go through a four-layer CNN with no padding, 3x3 kernels, stride 1, 32 output channels and ReLU activations. The height and width of the input to this CNN were both 8 larger than the target output (i.e. image) size to arrive at the target size (i.e. accommodating for the lack of padding). A final 1x1 convolutional layer transforms the output into 4 channels: 3 RGB channels for the means of the image components $\vxh_k$, and 1 for the logits used for the softmax operation to compute the reconstructed attention masks $\vmh_k$. For all experiments, the output component distribution was an independent pixel-wise Gaussian with fixed scales.

For the \ourawesomemodel experiments, the first "background" component scale was fixed at $\sigma_{bg} = 0.09$, and for the $K - 1$ remaining "foreground" components, the scale was fixed at $\sigma_{fg} = 0.11$. The loss weights were $\beta = 0.5$, $\gamma = 0.5$: 

For the component VAE experiments in Section.~\ref{S:exploiting_composition}, a single scale $\sigma = 0.05$ was used for all $K$ components, and we used $\beta = 0.5$, $\gamma = 0.25$.

\subsection{Attention network}
At the $k$th attention step, the attention network receives the concatenation of the input image $\vx$ and the current scope mask in log units, $\log \vs_k$, as input.

We used a standard U-Net \citep{ronneberger2015u} blueprint with five blocks each on the downsampling and upsampling paths (except for the \clevr experiments which used six blocks on each path). Each block consists of the following: a 3x3 bias-free convolution with stride 1, followed by instance normalisation \citep{ulyanov2016} with a learned bias term, followed by a ReLU activation, and finally downsampled or upsampled by a factor of 2 using nearest neighbour-resizing (no resizing occurs in the last block of each path).

Skip tensors are collected from each block in the downsampling path after the ReLU activation function. These are concatenated with input tensors along the upsampling blocks before the convolutional layer.

A 3-layer MLP serves as the non-skip connection between the downsampling and upsampling paths with its final output dimensionally matching that of the last skip tensor. The intermediate hidden layers were sized (128, 128). The input to the MLP is the last skip tensor collected from the downsampling path (after flattening). A ReLU activation is applied after all three output layers. The final output is then reshaped to match that of the last skip tensor, concatenated with it, and finally fed into the upsampling path.

Following the upsampling path, a final 1x1 convolution with stride 1 and a single output channel transforms the U-Net output into the logits for $\valpha_k$. Both $\log \valpha_k$ and $\log (1 - \valpha_k)$ are computed directly in log units from the logits (using the log softmax operation). Each are added to the current scope (also maintained in log units) $\log \vs_{k-1}$ to compute the next ($\log$) attention mask $\log \vm_k$ and next ($\log$) scope $\log \vs_k$, respectively. Also see Section ~\ref{S:monet} for equations describing the recurrent attention process.
\subsection{Optimisation}
All network weights were initialized by a truncated normal (see \citep{ioffe_2015}), and biases initialized to zero. All experiments were performed in TensorFlow \citep{Abadi_etal_2015}, and we used RMSProp for optimisation with a learning rate of $0.0001$, and a batch size of 64.

\section{Datasets}\label{S:datasets}
\begin{figure}[t!]
  \centering
  \includegraphics[width=0.9\linewidth]{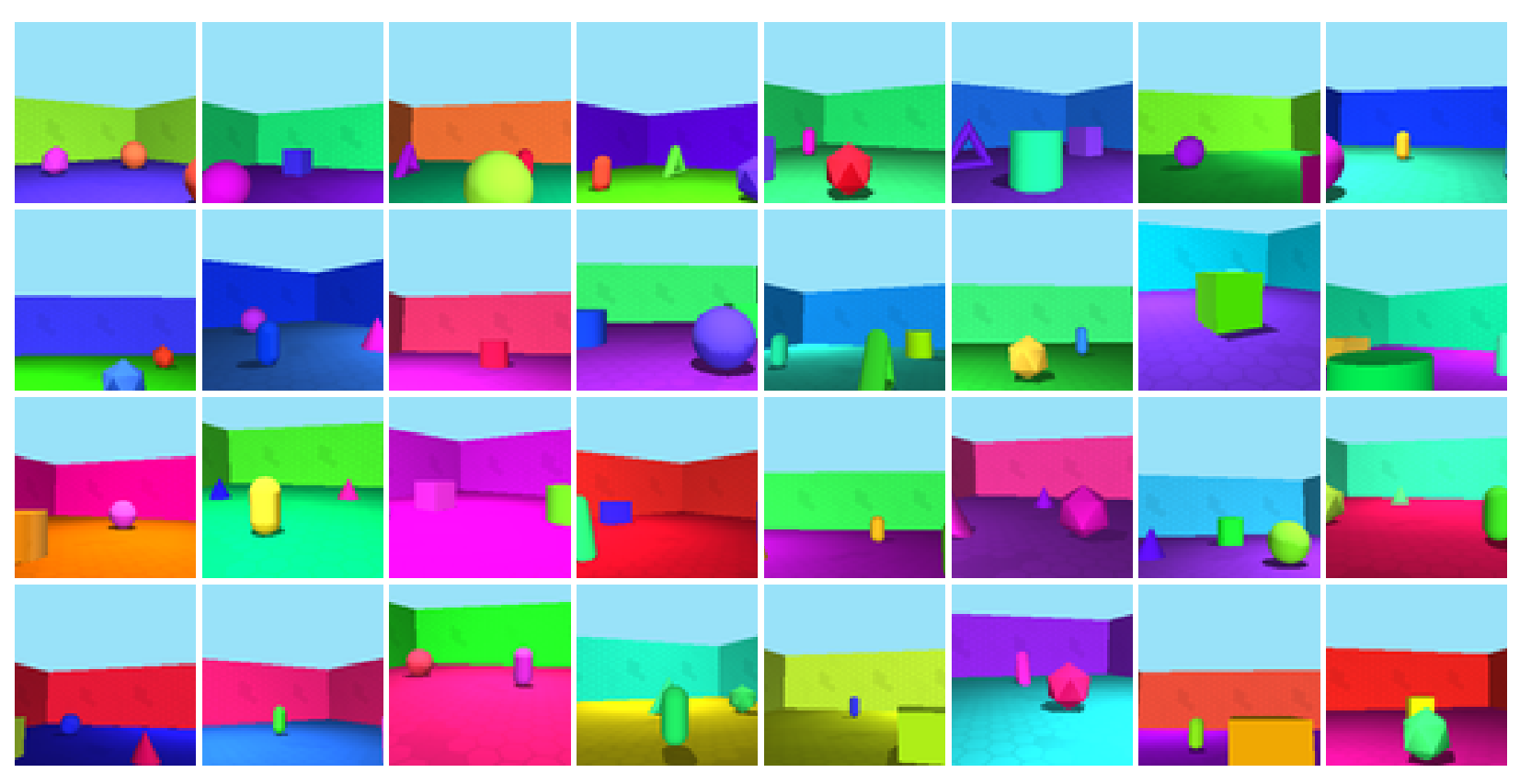}
  \caption{\textbf{The \oir dataset.} 32 random examples shown. See Section.~\ref{S:datasets} for more details.}
  \label{fig:objects_in_room_dataset}
\end{figure}

The \oir dataset was created using a Mujoco environment adapted from the Generative Query Networks datasets \citep{Eslami2018}. It consists of 64x64 RGB static scene images of a cubic room, with coloured walls, floors and a number of objects visible (see examples in Figure~\ref{fig:objects_in_room_dataset}). A blue sky is visible above the wall. The camera is randomly positioned on a ring inside the room, always facing towards the centre but randomly oriented vertically, uniformly in $(-25^\circ, 22^\circ)$. The wall, the floor and the objects are all coloured randomly, with colours each sampled uniformly in HSV colour-space, with H in $(0, 1)$, S in $(0.75, 1)$ and V always $1.0$. The objects are randomly sized, shaped (six different shapes), and arranged around the room(avoiding overlap). In the training dataset there are 3 objects in the room (making 1-3 objects visible).

The \multidsprites experiments used a dataset of 64x64 RGB images of 1-4 random coloured sprites. These were generated by sampling 1-4 images randomly from the 64x64 dSprites dataset \citep{dsprites17}, colourising the sprites with a uniform random RGB colour, and compositing those (with occlusion) onto a uniform random RGB background.

For the \clevr experiments we used images from the \clevr dataset \citep{Johnson2017clevr}. The standard images are 320x240, and we crop those at y-coordinates (29, 221), bottom and top, and at x-coordinates (64, 256) left and right (creating a 192x192 image). We then resize that crop using bilinear interpolation to 128x128.

\end{appendices}